%% file: main.tex
\documentclass[letterpaper, 10 pt, conference]{ieeeconf}  

\IEEEoverridecommandlockouts                             
                                                          
\usepackage{cite}
\usepackage{amsmath,amssymb,amsfonts}
\usepackage{algorithmic}
\usepackage{textcomp}
\usepackage{xcolor}
\usepackage[textsize=footnotesize]{todonotes}
\usepackage{verbatim}
\usepackage[nolist]{acronym}
\usepackage{dirtytalk}
\usepackage{subcaption}
\usepackage{graphicx}
\usepackage{comment}

\usepackage{theorem}
\usepackage{morefloats}
\usepackage{makeidx}

\newcommand{\editing[1]}{\textcolor{black}{#1}}

\title{Virtual Reality for Robots}
\author{}
\date{June 2019}

\input{acronyms.tex}

\input{defs.tex}

\author{Markku Suomalainen\textsuperscript{1} \hspace{5mm} Alexandra Q.\ Nilles\textsuperscript{2}\hspace{5mm} Steven M.\ LaValle\textsuperscript{1}
\thanks{This work was supported by Business Finland project HUMORcc 6926/31/2018, Academy of Finland project PERCEPT, 322637 and US National Science Foundation grants 035345, 1328018.}
\thanks{\textsuperscript{1}M.\ Suomalainen and S.\ M.\  LaValle are with Center of Ubiquitous Computing, Faculty of Information Technology and Electrical Engineering, University of Oulu, Finland. {\tt\small firstname.lastname@oulu.fi} \newline
\hspace*{1.1em}\textsuperscript{2}A.\ Nilles is with the Department of Computer Science, University of Illinois at Urbana-Champaign
, Illinois, USA. {\tt\small nilles2@illinois.edu} 
}} 

\begin{document}

\maketitle

\begin{abstract}
This paper applies the principles of Virtual Reality (VR) to robots, rather than living organisms.  A simulator, of either physical states or information states, renders outputs to custom displays that fool the robot's sensors. This enables a robot to experience a combination of real and virtual sensor inputs, combining the efficiency of simulation and the benefits of real world sensor inputs. Thus, the robot can be taken through targeted experiences that are more realistic than pure simulation, yet more feasible and controllable than pure real-world experiences. We define two distinctive methods for applying VR to robots, namely \textit{black box} and \textit{white box}; based on these methods we identify potential applications, such as testing and verification procedures that are better than simulation, the study
of spoofing attacks and anti-spoofing techniques, and sample generation for machine learning. A general mathematical framework is presented, along with a simple experiment, detailed examples, and discussion of the implications.
\end{abstract}

\section{Introduction}\label{sec:intro}
Virtual reality involves creating and maintaining an illusion that causes an organism to have a targeted perceptual experience.  This could be considered as {\em perception engineering} because the end result is perception, rather than a physical device.  We then wonder, what would it mean to analogously engineer a perceptual experience for a robot?

\begin{figure}[t]
        \centering
        \begin{subfigure}[b]{0.185\textwidth}
            \centering
            \includegraphics[width=\textwidth]{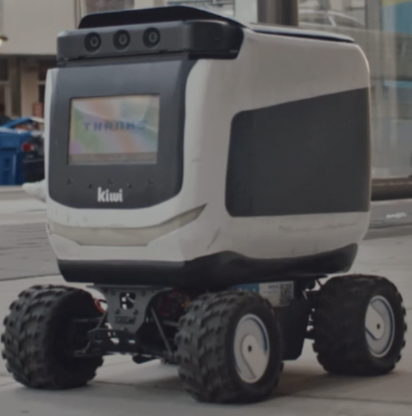}
            \caption{}
            \label{fig:kiwi}
        \end{subfigure}
        \begin{subfigure}[b]{0.235\textwidth}
            \centering
            \includegraphics[width=\textwidth]{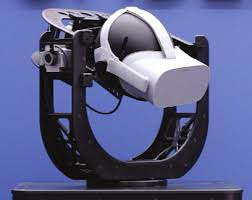}
            \caption{}
            \label{fig:buddy_vr}
        \end{subfigure}
        \caption{\small{(a) A Kiwibot food delivery robot. (b) A robot in full VR: An Optofidelity Buddy 3  wearing an Oculus Go headset.}}
        \label{fig:failmodes}
        \vspace{-0.5cm}
\end{figure}

Imagine a mobile robot, such as the food delivery Kiwibot in Fig.~\ref{fig:kiwi}. Assume it gets input data from three sensors: GPS, a front-facing camera, and wheel encoders. When the robot moves in cities and among people, it can easily lose GPS signal, get kicked or stuck, or even get kidnapped. These scenarios raise a question: What is a feasible but realistic method for replicating these scenarios during development and testing? Some scenarios, such as getting stuck, can be difficult to realistically simulate, whereas others such as crowd navigation are tedious or expensive to manage in real experiments. Moreover, issues such as system integration or systematic sensor errors cannot be completely replicated in simulation; thus, we would like to prioritize interaction with the actual, physical robot. \editing[How can we blend simulation and real-world measurements to achieve principled, systematic interaction with the physical robot?] How would such an interaction affect testing, reverse engineering, and verification of robotic systems? \editing[In fact, many robotics laboratories are currently developing infrastructure that provides a mixture of real and virtual inputs to robots for these purposes. By studying this infrastructure as a first-class subject, we aim to provide guiding principles for the development of these systems and help avoid over-engineering and resource waste.]

Virtual Reality (VR) uses simulation and displays to trick humans and other organisms into believing they are having a perceptual experience that is different from reality.  This experience is usually interactive and carefully crafted, such as games exploring exotic worlds. Other organisms, such as ants \cite{harvey2009intracellular}, pill bugs \cite{nagaya2017anomalous}, or \textit{Drosophila}, interact with artificial worlds so that scientists can study their creation of neural place and grid cells or other neurological phenomena. This paper explores a natural question:  {\em What would happen if the living organisms are replaced by a robot?}  This immediately raises additional questions. What new approaches would this enable? How would this form of VR be generally defined and achieved, well beyond what might be imagined from Fig.~\ref{fig:buddy_vr}? How would VR be successfully achieved in this context?  We call this class of problems {\em Virtual Reality for Robots (VRR)}.

\editing[In this paper we consider that all VRR applications can be considered as a continuum between two options: \textit{black box}, where we have no knowledge about the robot's internal algorithms, and \textit{white box}, where the internal algorithms are known; these concepts will be properly defined in Section~\ref{sec:general}. We ground this classification with] distinct applications for VRR, which revolve around the ability to create mixtures of real and virtual inputs:

\begin{enumerate}
    \item \emph{Testing:} The designers of a robot system want to evaluate its performance in environments that are difficult to reach or construct, or infeasible to simulate.  In this case, VRR provides a hybrid approach of real and virtual worlds that is more realistic than simulation but more accessible  than a normal deployment. \editing[Testing is largely a white-box application: we have a model of the robot's ``brain,'' and use VRR to ensure its accuracy in the face of challenging environments.]
    \item \emph{Spoofing:} As security and delivery robots increase in popularity, their protection against sensor spoofing attacks must be considered. VRR could be used to intentionally create sensor spoofing attacks on the actual sensors, and to study the robustness of the robot system. \editing[In this case, developing defenses will take a white-box approach, whereas developing attacks may take a black-box approach unless the adversaries have sufficient knowledge to mount a white-box attack.]
    \item \emph{Reverse Engineering:} To reverse engineer the design of an unfamiliar \editing[black-box] robot, VRR can put it through carefully designed, contrived scenarios. The robot receives an adaptive series of tests to determine its inner decision making strategies.
    \item \emph{Learning:} In a machine learning context, precious new data can be generated by measuring how the actual robot responds to numerous hybrid scenarios, addressing the gap between simulation and reality ({\em sim-to-real}; see e.g.~\cite{peng2018sim}). \editing[In this case the learned parts of the robot system are black boxes, but given our knowledge of the rest of the system we can still reason about the set of internal states of the robot as well as the necessary distribution and resolution of provided data.]
\end{enumerate}

\editing[The main contribution of this paper is the definition and formalization of an emerging field, providing both practical guidelines and a mathematical framework that can be used to design efficient VRR experiences for various scenarios]. Section \ref{sec:thought} introduces background on virtual reality concepts and their application to robotic systems. Section \ref{sec:general} provides a novel mathematical framework that captures the essence of VR for living organisms, but characterizes it in the general context of robots and provides some initial results on the resource requirements of VRR systems. This framework builds upon the usual state spaces (with configurations and environments), sensor mappings, and state transition mappings; we then introduce VR-specific notions such as a virtual world generator, renderers, and displays, which are used to fool the robot's sensors. VRR displays do not necessarily resemble a display or video screen in the usual sense; instead, each is custom designed to spoof a particular sensor, techniques for which will be explained in Section~\ref{sec:dd}. Section~\ref{sec:simren} discusses virtual world and rendering challenges.  Section~\ref{sec:experiment} presents a simple mobile robot demonstration, and  Section~\ref{sec:con} concludes by assessing the differences between VR and VRR, and speculating on the implications of this work.

\subsection*{Related work}
\editing[There are many novel systems that fall under our definition of VRR, meaning a mixture of real and virtual sensory inputs to a robot system. The driving forces behind these systems include the expense or difficulty of full-scale realistic experiments, especially in the multi-robot setting. Zhang et al.~\cite{zhang2019vr} developed ``VR-Goggles for Robots,'' demonstrating a virtual ``display'' for visual data that emulates simulated training data and improves performance of a learning-based system. Guerra et al.~\cite{guerra2019flightgoggles} demonstrated providing photorealistic simulated video and real dynamics for drone development. H{\"o}nig et al.~\cite{honig2015mixed} presented Mixed Reality for robots and demonstrate compelling use cases for the study of multi-robot systems. Shell and O'Kane \cite{shell2019reality} consider the theoretical requirements of constructing ``illusions'' for robots, mainly for multi-robot system applications, and provide interesting insights into the overhead of such systems with a Robotarium demonstration. While Shell and O'Kane do not use VR terminology, the ideas map well to our discussion of using VRR to drive a robot system into a desired internal state. ]

\textit{Spoofing} literature is also related, which considers adversarial attacks against sensor systems. Many attacks are against biometric security systems, such as face \cite{erdogmus2013spoofing}, fingerprint \cite{wild2016robust} or speech \cite{wu2015spoofing} recognition systems. There also exists mathemetical analysis on when spoofing is feasible \cite{zhang2016functional,zhang2018strategies}. These works, however, aim at fooling a classifier, whereas VRR is meant for continuous fooling of entire sensing and information processing subsystems. Other recent works show that MEMS sensors such as \acp{imu} can be distracted \cite{son2015rocking} and even controlled \cite{trippel2017walnut} with external amplitude-modulated noise. LiDAR, a key component in many autonomous cars, has been shown to be susceptible to spoofing attacks \cite{shin2017illusion}. GPS is not immune either, and GPS-based capture of autonomous vehicles is a major concern \cite{kerns2014unmanned}. Anti-spoofing methods for drones have also been proposed, by observing whether the combination of sensor inputs obeys the laws of physics \cite{davidson2016controlling}. These examples help to enable VRR (we can provide controllable input to real sensors) and further motivate it. To study and counter spoofing attacks, the concept of VRR must be well defined and understood. 


\section{Experiencing Virtual Reality}\label{sec:thought}

In this section we give background and definitions on \ac{vr} for humans
and other living organisms.
Then, we will show the connections to VRR, and define \textit{full \ac{vr}} and \textit{partial \ac{vr}}. 

\subsection{How living organisms experience virtual reality}

\begin{figure}[t]
\centering
\includegraphics[width=\columnwidth]{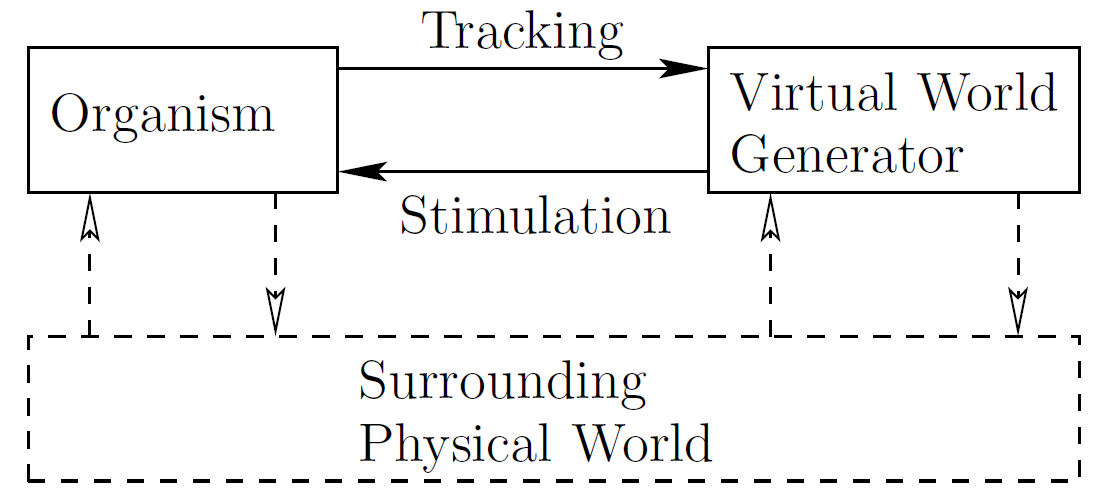}
\caption{\small{Information flows in VR for organisms. The organism interacts directly with a virtual world generator, which tracks the organism and has renderers that calculate display outputs based on the tracked state and the ``simulated world." Rendering on displays causes stimulation of the organism. The surrounding physical world is still present in the diagram because the organism can still potentially sense the ``real world" even when immersed in VR.}}
\label{fig:universe}
\vspace{-0.3cm}
\end{figure}

VR for living organisms can be defined as ``inducing targeted behavior in an organism by using artificial sensory stimulation, while the organism has little or no awareness of the interference" \cite{Lav19}. Creating
``targeted behavior" requires tracking the organism and rendering the virtual world accordingly, as shown in Fig.~\ref{fig:universe}. ``Awareness of the interference" refers to the phenomenon of \textit{presence} \cite{sanchez2005presence}, and is an important criterion of a successful human \ac{vr} experience. Interestingly, according to ``poison theory" \cite{laviola2000discussion} the failure of human sensor fusion to accept a credible situation is proposed as one of main reasons causing cybersickness; in such a situation being poisoned is a possible cause, and thus vomiting is a reasonable reaction. Similarly, if a robot expects to be spoofed, such a non-viable set of sensor inputs could warn the system of this possibility, analogously as proposed in \cite{choi2018detecting}. 





Next, consider ``artificial sensory stimulation." Fig.~\ref{fig:universe}
demonstrates how a \ac{vr} experience is generated; note that
``stimulation" is not limited to visual stimuli but includes \textit{all} possible sensing modalities. Besides vision, common sensing modalities include audio and tactile feedback, proprioception (through e.g., treadmills), olfaction \cite{salminen2018olfactory}, and even
\ac{ems} \cite{lopes2017providing}. Also, fooling vision does not necessarily require a \ac{hmd}. For example, the ``visual odometry" of honeybees can be influenced by changing the pattern of a tunnel they flew through \cite{srinivasan1996honeybee}. Also, for humans, {\em CAVE} systems (i.e., surrounded by screens) are considered VR \cite{cruz1993surround}. 

Creating artificial sensory stimulation that achieves the conditions of targeted behavior with little or no awareness of the interference is often a challenging task. Software must take advantage of the strengths of the hardware while hiding the flaws that would give away the interference.

If a VR system does not target all sensing modalities, we will call it \textit{partial VR}; let \textit{full VR} mean that all sensors are targeted, in which case the target of the VR experience has no knowledge or contact with the surrounding physical world. Biological sensor systems such as proprioception, the vestibular system, and temperature/pressure sensing in the skin are quite complex. Thus, achieving \textit{full \ac{vr}} on humans or other organisms is practically impossible; this would mean that the user does not feel the temperature in the room, or the pressure of the floor under the feet. Given their relative simplicity, full VR is feasible for robots, and would look like a system that could spoof every sensor onboard the robot. The main advantage would be integration and sensor testing, since the sensor input is completely controlled; however, a more interesting use case is mixing real and virtual inputs to find an optimal combination between realisticity and reproducibility. Thus, for the remainder of this paper, we consider mainly various levels of partial VRR. 


\subsection{How robots might experience virtual reality}



\editing[The primary motivations for VRR over simulation are a) allowing the robot to experience a mixture of real and virtual inputs and b) to test the real sensors instead of simulated ones.] We consider VRR by direct adaptation of how VR is currently defined for humans, \editing[with the exception that it is possible to bypass a subset of the robot's sensors to inject
virtual inputs directly into the algorithm]. The way humans experience VR, if used for robots, implies that the only way to affect the robot is to interact with its sensors through a \textit{display}, to which \textit{stimuli} are \textit{rendered} from a \textit{virtual world}, as shown in Fig.~\ref{fig:universe}. For the stimuli to be rendered correctly, the VRR system must also \textit{track} and possibly \textit{predict} the actions of the robot. Whereas perception engineering can be of use here, a major difference is the way robots ``perceive", which is mainly their sensor fusion and control systems. Thus, the challenges are different, which is further explored in Section~\ref{sec:general}.

While including the real sensor is better for system integration and testing, we acknowledge the difficulty of building a physical display for each sensor and also the results of previously mentioned works \cite{guerra2019flightgoggles,zhang2019vr,honig2015mixed} who achieved significant results while bypassing the physical sensors. The rest of this paper is written from the point of view of building real, physical displays, but much of our reasoning holds true for both methods.

Rendering is not limited to a visual display, but encompasses all the sensors being fooled, to be discussed in Section~\ref{sec:dd}. This provides the main strength of VRR: the ability to create a mixed world of \textit{real} and \textit{virtual} sensor inputs. For example, simulation of dynamics involving air and fluid flow, or slipping on granular materials, is still computationally intractable and thus prone to artifacts and assumptions in simulation. Also, light and sound reflections are still computationally prohibitive to simulate completely realistically.
Moreover, in simulators it is often assumed that sensors are perfect, or if noise is considered, then the injected noise is almost always assumed to be Gaussian. However, it is usually the case that the Gaussian assumption is inadequate. There may be unexpected systematic errors not captured in simulation, such as temperature dependence of sensors, or interference between electromagnetic or ultrasonic sensors. By retaining real sensors on the robot, VRR enables targeted experiments with results that can be more reasonably expected to transfer to deployment.

\section{General Mathematical Framework}\label{sec:general}

This section takes the considerations from Section \ref{sec:thought} as a starting point and both formalizes and generalizes them into a precise mathematical framework that captures VRR and extends naturally from typical robotics models. \editing[To provide VRR, the robot should be ``tricked" into executing a target behavior or plan, even though a subset of sensor may receive inputs from the real world and the robot may not traverse the workspace in the way that it ``believes."  The VRR problem is how to ensure that every sensor provides observations that the robot would expect based on execution of the target plan. By understanding the resource requirements of the scenario, the system and displays, VRR systems can be designed efficiently, avoiding over-engineering. ]

Consider a generic robot that is equipped with sensors, actuators, and computation.  Let $X$ denote its standard {\em state space}, which could be its configuration space or more general phase space to include time derivatives.  Let $U$ denote its {\em action space}, which corresponds to the set of commands that can be given to the robot. The robot has one or more sensors, each modeled by a sensor mapping; the most common form of this mapping is $h: X \rightarrow Y$, but in general, $X$ could be replaced by a space that includes time, state histories, or unpredictable disturbances \cite{lavalle2012sensing}.  Each $y = h(x)$ is called a {\em sensor observation}.

A {\em state transition equation} $x' = f(x,u)$ determines the effect of the action $u \in U$ when applied at state $x \in X$, resulting in a new state $x' \in X$.  The system feedback loop could occur over discrete time at fixed intervals, be event driven, or any other common possibility.  

The robot selects an action $u$ according to a {\em plan}, which has the form $\pi: \I \rightarrow U$, in which $\I$ is an {\em information space}, defined for robots in \cite{Lav06}, but derived from \cite{BasOls95,VonMor44}.  They become belief spaces in a Bayesian setting \cite{KaeLitCas98}, but are not necessarily restricted to probabilistic models here.  The action $u = \pi(\eta)$ is chosen based on an {\em information state} or {\em I-state} $\eta$, which is derived (a mapping) from initial conditions, the sensor observation history, the action history, and possibly the state transition equation and sensor mappings.  A common example is that in a Bayesian setting, $\eta$ corresponds to a posterior pdf that takes into account all models and existing information.  As another example, $\eta$ could simply be the most recent sensor observation, $y$, resulting in pure sensor feedback.  Note that there is no direct access to the state $x$ at any time (unless a powerful enough sensor can measure it, which is unlikely in practice).



Let $D$ denote a {\em display output space}, in which a particular {\em display output} is denoted by $d \in D$.  A display is associated with a sensor, implying a relationship in which the display output $d$ causes a targeted observation $y$.  Thus, there is a function $\sigma : D \rightarrow Y$, called the {\em spoof mapping}.  More generally, the spoof mapping may depend on state; an ideal display would not have this dependency, but in reality issues such as outside lighting having an effect on a screen must be considered. This yields ${\sigma: D \times X \rightarrow Y}$.\footnote{Of course, the display is embedded in the physical world. We use this notation for the spoof mapping for clarity, implicitly defining $X$ as ``the rest of the world.'' Ideally, the display would not alter configuration space obstacles; in reality this will depend on the design of the display.} 

How does the display know what to output?  For human-based VR, a \textit{\ac{vwg}} (see Fig.~\ref{fig:universe}) maintains a virtual world, which is then rendered to displays according to the tracked configuration of the human.  The same idea is needed for VRR.  Let $S$ denote the {\em virtual state space}. Note that $S$ and $X$ could be the same or vastly different.  Each display uses the state $s \in S$ to determine the output through a {\em rendering mapping}, $r : S \rightarrow D$.  Thus, $d = r(s)$ is the rendered output to the display when the \ac{vwg} is in state $s$.  

More generally, the display output might depend on both the physical state $x$ and the virtual state $s$.  In this case, the rendering mapping is $r : S \times X \rightarrow D$ and $d = r(s,x)$.  This analogously happens in human-based VR, in which we must know where the user is looking (equivalent to $x$) and what in the virtual world needs to be rendered (equivalent to $s$).  The implementation of $r$ might then require a tracking system to estimate $x$ (analogous to VR head tracking \cite{LavYerKatAnt14,WelFox02}); in this paper we will assume that the VRR system includes sufficiently accurate tracking.

Creation of the Virtual World Generator can be defined between two extremes, depending on our knowledge of the robotic system:
\begin{enumerate}
\item {\em Black-box system:} If we have no knowledge of the internal algorithms of the robot, then the \ac{vwg} should maintain a complete and perfect virtual state space, mimicking the behavior of the real world with sufficient fidelity that the appropriate display outputs can always be determined. 
\item {\em White-box system:} If we know the internal algorithms of the robot, the \ac{vwg} can directly induce the transitions of I-states inside of the robot with an incomplete or imperfect virtual state space; the \ac{vwg} need not maintain a high-fidelity artificial world.
\end{enumerate}

The first choice is appropriate when we do not have direct access to the I-states. This is the common situation in human-based VR, in which it is impossible to measure or understand the brain's I-state (all relevant neural activity). The second choice is available for VRR but not human-based VR because we might have access to the robot's design. Whereas the extreme of white box may not provide interesting use cases, the concept can be extended to cases where we have \textit{partial} knowledge, or assumption, about which features are important for the robot's internal algorithms. The implications of this are quite powerful. For example, if we know that the I-state completely ignores one sensor, then there is no need to design a display for it. Consider a range finder that is only used to report an obstacle within a certain threshold. For such coarsely quantized sensor observations, the display resolution could be significantly lowered. The \ac{vwg} may more resemble a simple state machine that emits the correct displays rather than a physics simulator. 

\subsection{Black-box systems}

Consider the case of a robot where we know its sensors and their mappings, and can track the robot's actions, but know nothing of its algorithms or internal state. The \ac{vwg} ideally would maintain a physically plausible world, of sufficient fidelity that no sensor would be able to detect the presence of the artificial display. Precise information about the robot's supposed configuration in virtual and real environments may be maintained by a tracking system.

The sensor mapping implies a \emph{sensor preimage} for each sensor reading $y \in Y$, defined as
\begin{equation}
h^{-1}(y) = \{x \in X \mid y = h(x)\}.
\end{equation}
Since sensors usually are many-to-one mappings, $h^{-1}(y)$ could be a large subset of $X$.
For example, a proximity sensor that returns \textsc{True} if the
sensor is within five centimeters of a wall, and \textsc{False} otherwise, $h(x) = \textsc{True}$ would induce a preimage that correspond to all states that put the sensor within five centimeters of a wall.


In general, the collection of sensor preimages for all possible sensor readings forms a partition of $X$. For a given sensor mapping $h$, let the partition be denoted $\Pi(h)$. The sets in $\Pi(h)$ should be thought of as equivalence classes, because for any two $x_i, x_j$ in the same sensor preimage, $h(x_i) = h(x_j)$ and the states are indistinguishable with that single sensor.

This has implications for the necessary resolution of the virtual state space. Whereas $X$ may be continuous, $S$ can be formed by discretizing $X$ as the common refinement of all partitions $\Pi(h)$ for all sensors. Within each element of $S$, all corresponding states $x$ will then be guaranteed to produce the same sensor readings. Two challenges remain: First, some sensors (such as cameras) have such fine-grained mappings that this approach would be prohibitively complex. Second, a computational burden is added of needing to compute whether the robot will transition between states in $S$, based on its state transition model in $X$. However, especially for robots with simple sensing modalities (such as swarm robots or micro-scale applications), this approach can dramatically simplify the corresponding VRR system.

\subsection{White-box systems}\label{sec:white}

For the case of a white-box robot, we formalize how to use a VRR system to induce specific targeted behaviors in a derived information space. For example, a vacuuming robot may have I-states corresponding to cleaning in free space, cleaning along a wall, and finding a charger, with reactive transition rules between these states that we wish to test. The I-states of the robot may correspond to a set of possible physical states, a belief distribution over the physical state space, or something more abstract. In this paper, we show that for the case where the robot's internal decisionmaking algorithms can be represented by a deterministic finite automaton (DFA), it can be algorithmically practical to use a VRR system to induce target internal states. We see this as a first, simple, theoretical result for VRR with implications for the required complexity of VRR systems.

If we know the initial I-state, then inducing specific behaviors involves a straightforward search for a sequence of sensor values that drive the robot to the target I-state. The case in which we do not know the initial I-state is more interesting because it allows for analysis of behavior over all possible initial conditions, important for robustness testing in cases where we want a robot to react the same way to certain stimuli regardless of prior state. This approach is also a first step toward strategies for using VRR to reverse engineer an unknown robot. 

\vspace{2mm}
\noindent {\bf Proposition:} {\em A polynomial-time algorithm in $|\I|$ and $|Y|$ exists that computes a sequence of targeted sensor observations that will drive the robot from any start I-state to any desired target I-state, or else decide that no such sequence exists.}

\noindent {\bf Proof:} 
Assume the robot has a finite discrete I-space $\I$, finite discrete action space $U$, and finite discrete observation space $Y$ for every sensor.  Assume that the plan $\pi : \I \rightarrow U$ is known and deterministic.  Assume an event-based transition model, in which the robot performs an action and gets a sensor observation at each transition. In this case, the traversal of the information space can be mathematically treated as a deterministic finite automaton, with state space $\I$, input symbol space $Y$, and a state transition map that yields next I-states based on the current I-state, action according to $\pi$, and observation $y$. Each sensing history can be thought of as an input string that drives the robot into a specific I-state. Thus the size of the DFA's state space is $|\I|$, and the size of the input symbol space is $|Y|$. Since the robot's action at each stage is completely determined by it's I-state, the actions can be thought of as output symbols and are disregarded in this case.

Under these assumptions, our problem of driving the robot from any start I-state to a given target I-state becomes the {\em synchronizing word} problem in automata theory, also related to part orientation problems \cite{goldberg1993orienting}. We seek a word in the input alphabet of a given DFA that will send any initial state of the DFA to one and the same state \cite{volkov2008synchronizing}. In our problem, this is equivalent to finding a sequence of sensor readings that would cause a robot with known structure and plan to be in a specified I-state after the sequence, regardless of the original I-state. For general $n$-state DFAs over a $k$-letter input alphabet, an algorithm exists to find the a synchronizing word in time $\mathcal{O}(n^3+kn^2)$, and the bound on the length of the respective word is $\mathcal{O}(n^3)$ \cite{eppstein1990reset}. Depending on the structure of the problem, these bounds can be improved significantly. \qed

\vspace{-1em}

\editing[We interpret this result to underscore the usefulness of well-defined, modular reasoning systems for robots. If the structure of the robot's internal decisionmaking apparatus is simple enough, it seems feasible to build VRR systems that allow for comprehensive system testing and verification. As robot internals become more complex, attempting to build out infrastructure for testing all possible internal states of the robot may be a somewhat hopeless endeavour.]

Another important remark is that the above is an ``open-loop" approach to inducing targeted behaviors in a known robot; we do not observe the actions that the robot takes or use them to estimate the current I-state. The problem of adaptively estimating the I-state is an interesting open problem. If we can control some or all inputs to a robot's sensors, then under what conditions can observing its resulting actions give us enough information to conclusively determine its I-state? Can the approach be extended to the case where we are attempting to infer the structure of the information space, or deciding between a few candidate information space models?

\section{Designing Displays}\label{sec:dd}

The mathematical framework of Section \ref{sec:general} is simple but abstract.  This section provides more details on  the structure of the spoof mapping $\sigma: D \rightarrow Y$ for particular sensor-display combinations.

\subsection{Displays for a camera}
\label{sec:disp_camera}

Suppose the robot uses a standard RGB camera, for example, with VGA resolution, global shutter, and standard lens.  A sensor observation $y = h(x)$ would correspond to a complete specification of an image captured by the camera.  The observation space $Y$ is large, containing $640 \times 480 \times 256 \times 3$ elements.  We want a display so that for any $y$ of interest, there exists a display output $d \in D$ for which $y = \sigma(d)$.

\begin{figure}
\begin{center}
\begin{tabular}{cc}
\includegraphics[width=3.0truecm]{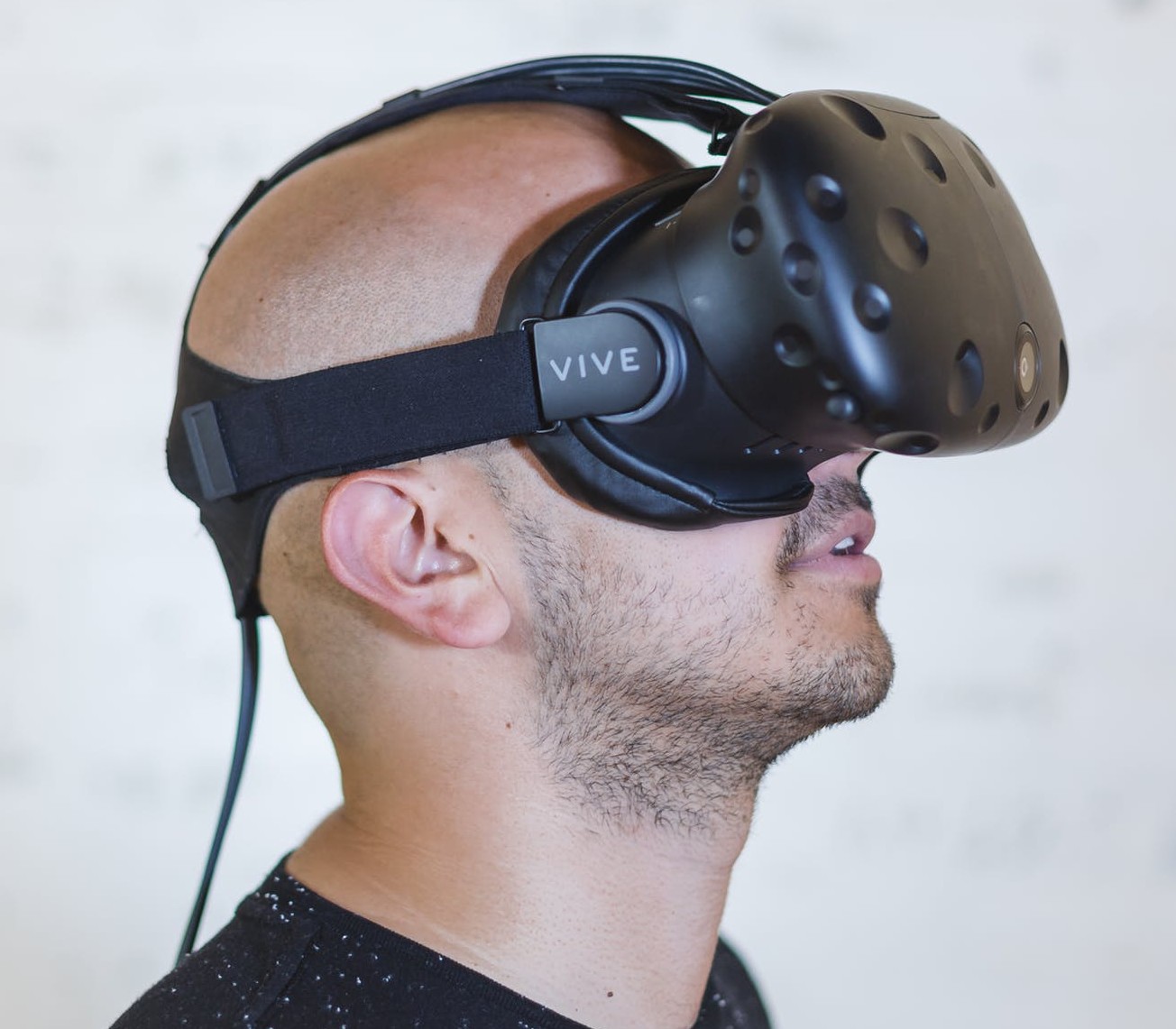} &
\includegraphics[width=4.0truecm]{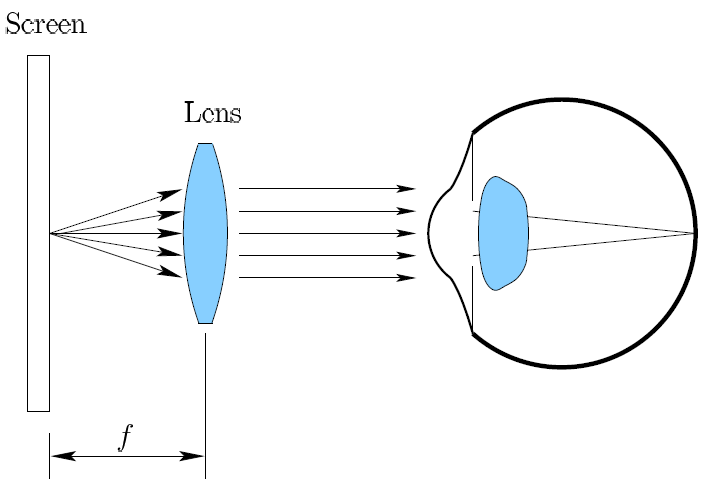} \\
(a) HMD & (b) Optical system 
\end{tabular}
\end{center}

\caption{\label{fig:hmd} \small{ (a) For human-based VR, the human vision system is spoofed by wearing a screen on the head and blocking external light. (b) A lens is placed between the screen and eye so that the display appears to be further away.}}
\vspace{-0.3cm}

\end{figure}

A common means to spoof a camera is by placing an ordinary RGB screen in front of it; this approach has been taken to spoof security systems in \cite{komulainen2013context}. For VRR, the approach is analogous to a human wearing an \ac{hmd}; see Figure \ref{fig:hmd}a.  Each display output $d$ would specify the eight-bit RGB values every pixel in a display image.  For a standard 1080p display panel, this would imply that the display output space $D$ contains precisely $1080 \times 1920 \times 256 \times 3$ values.

Placing a standard screen display over a camera causes several issues:  1) ambient light, if not blocked, also affects the camera input; 2)
A lens is required to make the camera focus correctly (see Figure \ref{fig:hmd}b), unless an exotic alternative, such as a light field display \cite{lanman2013near}, is used;
3) The camera might capture the rolling scanout of the display images, rather than a complete image, which suggests that vsync must be used to control the display and its frame rate should be significantly larger than that of the camera; 4) The resolution of the display should be significantly higher than that of the camera, in terms of pixels per degree; even though in theory matching resolution should suffice, the display is almost impossible to align perfectly to avoid severe quantization artifacts; 5) numerous other problems might affect performance, such as noise, limited dynamic range of the display, and optical aberrations. 

\editing[Finally, when spoofing white box robots, we may use assumptions about the robot's internal image processing algorithms to avoid full image generation by displaying only the features detected by popular computer vision algorithms. An example of such a spoofing was done by Su et al.~\cite{su2019one}, who managed to fool deep neural network classifier in 16.04\% of the ImageNet dataset images by altering only one of the 227x227 pixels.] 



\subsection{Displays for contact or proximity sensors}

Next, consider simple sensors for which $Y = \{0,1\}$, such as a mechanical contact or bump sensor.  In one mode, $y=0$, there is no contact.  In the other mode, $y=1$, the bumper is pressed and contact is made.  In this case, the display needs only to press the bumper to spoof the robot, which can be accomplished by a mechanical attachment.  In this case, $D = \{0,1\}$, and the spoof mapping takes an obvious form: $y = \sigma(d) = d$.  Thus, a ``display'' in this case merely smacks the contact sensor so that it reports contact!  The situation is similar for a typical proximity sensor.  If proximity is detected by a simple infrared detector, then an object needs to be placed into its field of view to report detection.  The set $D$ and mapping $\sigma$ remain the same as for the contact sensor. Naturally this interference must occur at desired intervals to create a ``virtual world" for the robot. Thus, the VWG must maintain information from which occlusions are rendered as appropriate to make the robot perform the targeted behavior. 

\subsection{Other display examples}


A display could be designed for any sensor aboard a robot. A force/torque sensor can be fooled by a more complicated smacker, such as a robot arm. A method for building a display for LiDAR was presented in \cite{shin2017illusion}, such that objects can appear closer than the display, or are even erased from the LiDAR.  Although we did not find existing work for other sorts of distance sensors, it is not difficult to imagine displays such as a fully sound-absorbing surface with a microphone and a loudspeaker for a sonar, or a system of adjustable mirrors for infrared. A \textit{dynamometer}, a sort of treadmill for measuring cars, can be used to fool wheel encoders while the mobile robot remains stationary. Whereas there is industrial interest for such a setup, in many use cases, such as detecting slippage or getting stuck, a more useful VRR design would be to spoof the other sensors and allow the real wheel encoder data; real surfaces with highly varying friction and potholes are typically infeasible to simulate.



\section{Virtual World and Rendering Challenges}\label{sec:simren}

Rendering in the classical computer graphics sense means generating an image from a model \cite{schroeder2004visualization}. To render into an \ac{hmd}, an additional element of \textit{tracking} is required because rendering depends on the device's location, thus combining the \textit{virtual} and \textit{real} worlds. Whereas rendering on an \ac{hmd} and a visual display meant to fool a robot's camera may sound similar at first thought, subtle differences must be taken into account. For \acp{hmd} and other screens meant for humans, displays have been optimized to ``fool" human eyes, for which there is an accepted notion of {\em normal vision}. However, because of the wide variety of possible cameras, it can be difficult to design a display that would fool any camera, due to challenges explained in Section~\ref{sec:disp_camera}. 

Rendering and virtual world models can be engineered for other sensors, and the corresponding displays will have unique tracking challenges.  Consider human-based VR.  Whereas audio rendering requires only head tracking, for haptic rendering all relevant degrees of freedom must be tracked. For VRR, the challenges regarding the virtual world and tracking are similar but broader. All parts of the robot that contain sensors to be spoofed may need to be tracked in the real world. Moreover, the idea of \ac{vwg} and rendering to a haptic display or an IMU is a concept that must be properly defined. Possible delays in rendering (smacking) must be considered. A display for an IMU would be more complicated, even though spoofing literature has shown that it is not impossible \cite{trippel2017walnut}.

Finally, any knowledge of the robot's internal algorithms can be used to simplify the complexity of the \ac{vwg}, as explained in Section~\ref{sec:white}. However, in many modern algorithms, the robot's internal state space may be intractably large. Modular design or state-space coarsening should be explored to make testing these systems more feasible.

\section{A Simple Proof of Concept}
\label{sec:experiment}

\begin{figure}[tb]
\centering
\includegraphics[width=0.8\columnwidth]{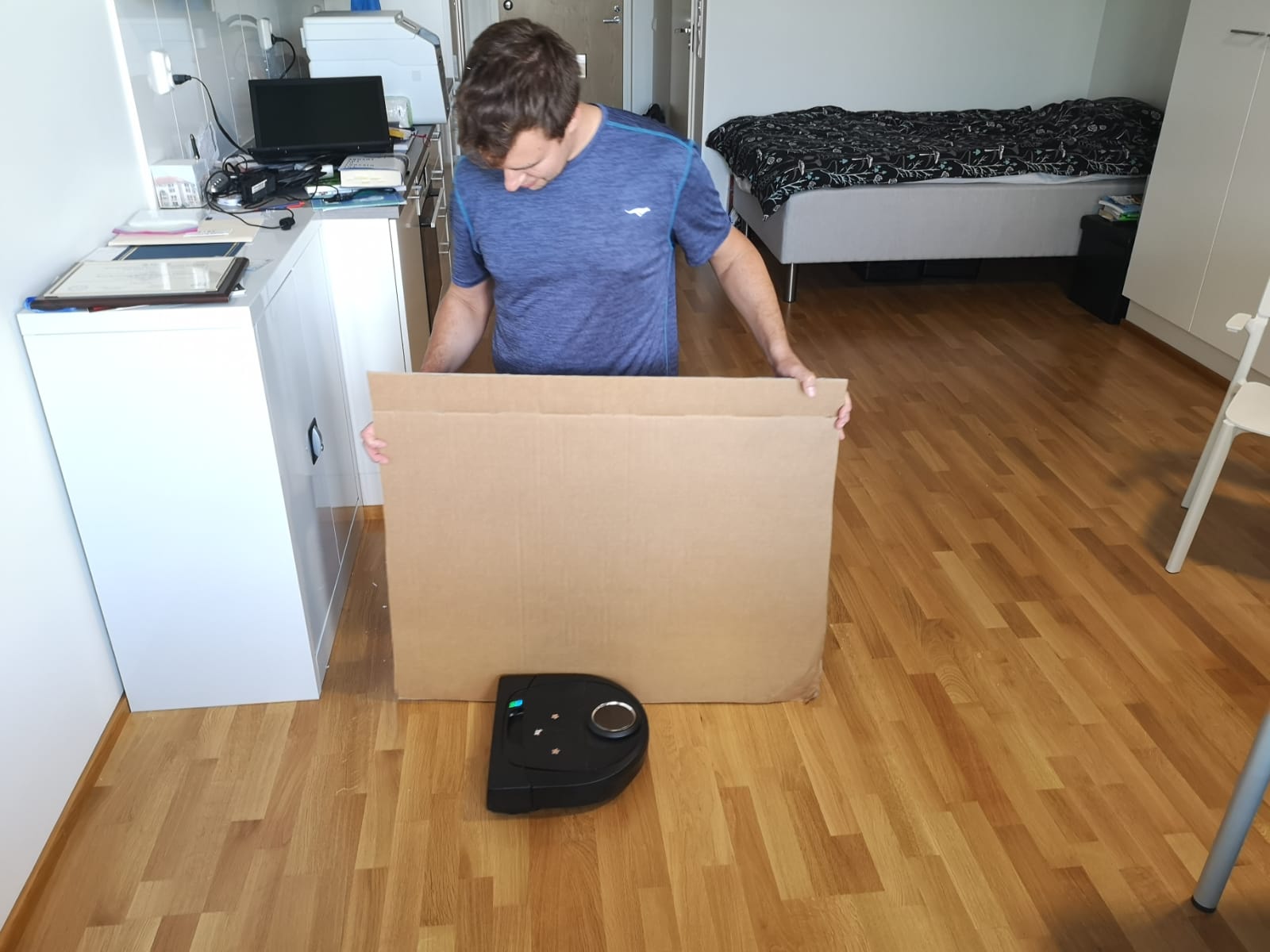}
\caption{\small{A human (Markku) with cardboard, acting as a \ac{vwg}, haptic renderer and display for a Neato vacuum cleaner robot.} }
\label{fig:neato_vr}
\vspace{-0.5cm}
\end{figure}

\begin{figure}[tb]
        \centering
        \vspace{4.0mm}
        \begin{subfigure}[b]{0.193\textwidth}
            \centering
            \includegraphics[width=\textwidth,trim={5cm 0 5cm 0},clip]{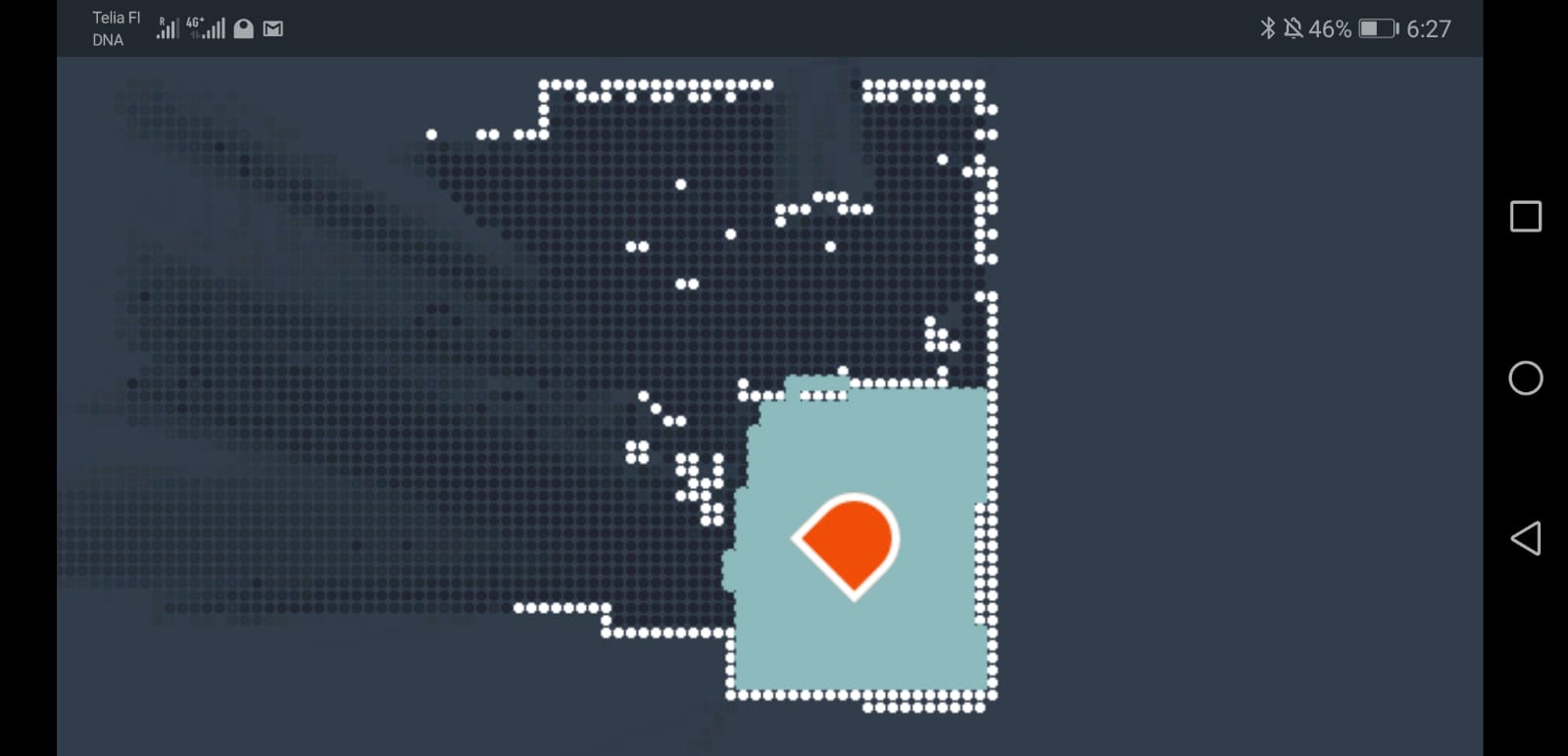}
            \caption{}
            \label{fig:neato_map1}
        \end{subfigure}
        \begin{subfigure}[b]{0.225\textwidth}
            \centering
            \includegraphics[trim={1cm 0 1cm 0},clip,width=\textwidth]{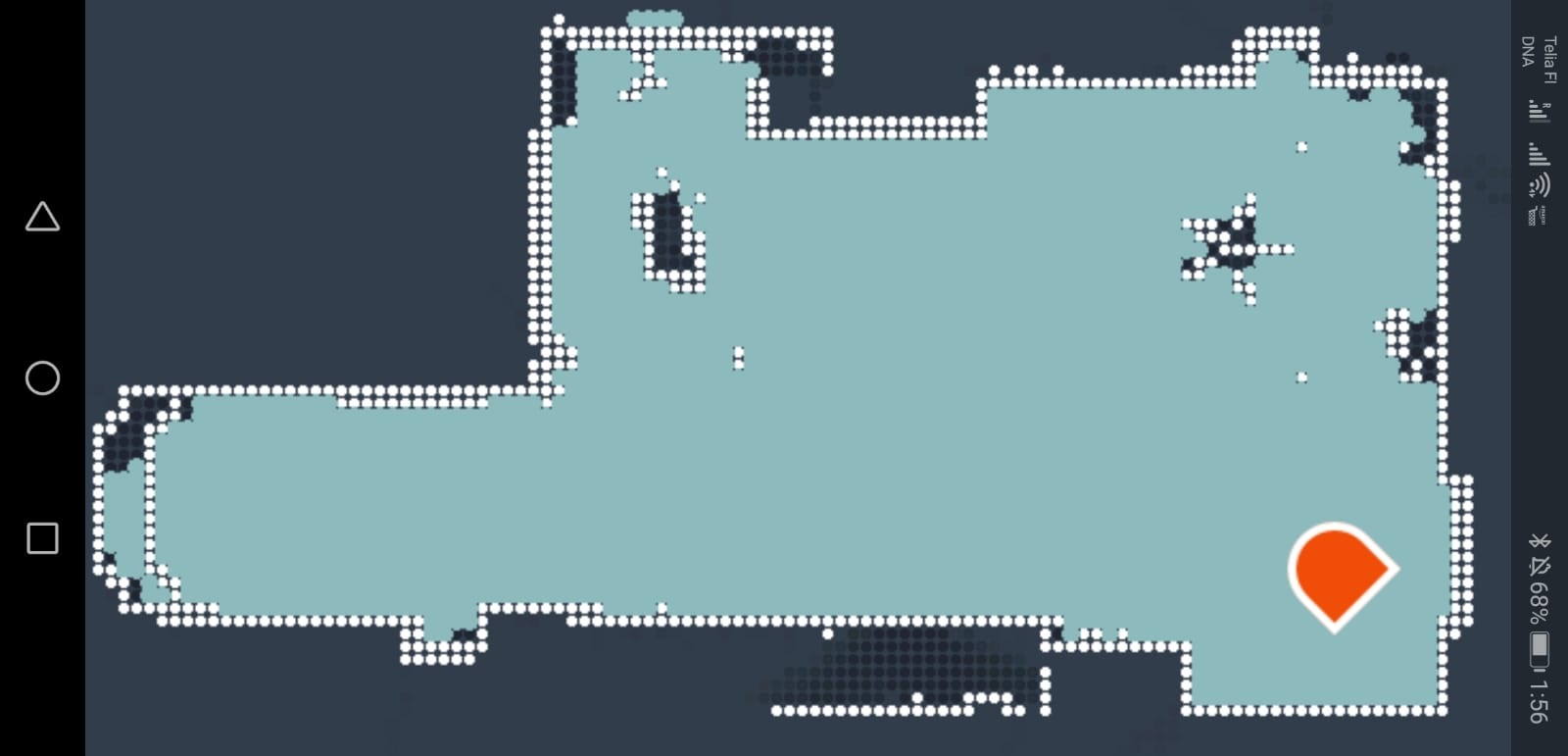}
            \caption{}
            \label{fig:neato_map2}
        \end{subfigure}
        \caption{\small{Two maps created by the Neato robot. (a) is the map where the robot thinks it cannot pass any further, and (b) is the full room.} }
        \label{fig:neato_maps}
        \vspace{-0.3cm}

\end{figure}

We performed a demonstration of how a vacuum cleaner robot, Neato Botvac D5, can be fooled to think it is in a smaller passable area than it really is. \editing[We take the Neato to be a black-box system, where we know it uses a bump sensor and a range finder, and we wish to reverse-engineer the robot to see if obstacles present only to the bump sensor are enough to constrain the motion of the robot.] The setup is shown in Fig.~\ref{fig:neato_vr}. \editing[The human, who acts as a virtual world generator, holds a piece of cardboard, a haptic display. The plan is to render stimuli to the robot's bump sensors so that the robot believes it cannot move beyond the designated area]. However, everything else (dynamics and obstacles, such as carpets and thresholds) is real, thus creating a partial VR environment for the robot. In Fig.~\ref{fig:neato_maps} are two maps created by the robot, where in (a) the haptic display is used, and in (b) the robot is free to use the whole room. Interestingly, we also observed that this haptic display is insufficient to completely fool the range finder, but those observed areas of the room were deemed impassable, thereby demonstrating the concept of partial VR being useful for inducing behaviors. 

\section{Conclusion}\label{sec:con}

This paper has introduced the notion of virtual reality for robots (VRR), by drawing parallels between VR for humans (and other organisms) and spoofing a robot's sensors.  This has led to a mathematical framework that contains general and formalized notions of displays, rendering, and VWG. Our definitions are directly interleaved with standard notions from robotics, including state spaces, actions, sensor mappings, state transitions, and information states. Using this framework, we identify several interesting open problems for further research, and guiding principles for systematic testing and experimentation. The engineering challenges involved are worth pursuing because of the enormous potential for applications in reliability testing, reverse engineering, security, and machine learning, as explained in Section \ref{sec:intro}. Tools and best practices for VRR are especially needed for creating samples for machine learning algorithms, optimizing the mixture of real and virtual inputs to maximize realism and minimize time and resource investments; VRR also allows controlled testing or tuning of learning systems in partial VR. 


We expect significant future work to emerge in both VR and robotics by leveraging the parallels and distinctions made in this paper.  Interesting VRR questions are inspired by human-based VR, and vice versa.  For example, most ``information" that would correspond to activation of photoreceptors on the retina is discarded or compressed by the ganglion, amacrine, horizontal, and bipolar cells before neural impulses are passed to the brain along the optic nerve \cite{Mat08}; this is analogous to the calculation of information states in a robot. Furthermore, photoreceptor density, sensitivity, and activity rates vary substantially along the retina.  Such understanding has motivated techniques such as {\em foveated rendering} for human-based VR \cite{guenter2012foveated}, and leads to questions such as how knowledge about sensor limitations and sensor fusion methods can be exploited to facilitate VRR solutions.  Likewise, the ability to completely know the inner workings of a robot may offer insights into the improvement of human-based VR, for which sensory systems, perception, and physiological effects are not fully understood (we did not engineer ourselves!).


\bibliography{references}
\end{document}

%% file: acronyms.tex
\newacro{vr}[VR]{Virtual Reality}
\newacro{hmd}[HMD]{Head-Mounted Display}
\newacro{ems}[EMS]{Electrical Muscle Stimulation}
\newacro{imu}[IMU]{Inertial Measurement Unit}
\newacro{bci}[BCI]{Brain-Computer Interface}
\newacro{vwg}[VWG]{Virtual World Generator}

%% file: defs.tex
\def\I{{\cal I}}

\def\qq1{{q_1}}

\def\qkp1{{q_{k+1}}}
\def\qKp1{{q_{F}}}

\def\qKp1{{q_{F}}}

\def\qxq1{{q_1}}

\def\uu1{{u_1}}

\def\xi{{x_i}}
\def\xkp1{{x_{k+1}}}
\def\xKp1{{x_{F}}}

\def\xq1{{x_1}}
\def\ukp1{{u_{k+1}}}

\def\atan2{\operatorname{atan2}}

\theoremstyle{plain}

\newcommand{\qed}{\hfill $\blacksquare$ \hfill \\}